\documentclass[runningheads]{llncs}
\usepackage{graphicx}
\usepackage{tikz}
\usepackage{comment}
\usepackage{amsmath,amssymb} % define this before the line numbering.
\usepackage{color}

\usepackage{multirow}
\usepackage{color}
\usepackage{graphicx}
\usepackage{amsmath}
\usepackage{algorithm}
\usepackage{algpseudocode}
\usepackage{subfigure}
\usepackage{booktabs}
\usepackage{enumitem}
\usepackage[accsupp]{axessibility}  % Improves PDF readability for those with disabilities.

\begin{document}
% \linenumbers
\pagestyle{headings}
\mainmatter
% \def\ECCVSubNumber{5098}  % Insert your submission number here

% Compact Neural Networks via Stacking Basic Units
% Compact Neural Networks with Basic Units
\title{Compact Neural Networks via Stacking Designed Basic Units} % Replace with your title

%\begin{comment}
% \titlerunning{ECCV-22 submission ID \ECCVSubNumber} 
% \authorrunning{ECCV-22 submission ID \ECCVSubNumber} 
% \author{Anonymous ECCV submission}
% \institute{Paper ID \ECCVSubNumber}
%\end{comment}
%******************

% \begin{comment}
\titlerunning{Compact Neural Networks via Stacking Designed Basic Units}
\author{Weichao LAN \and
Yiu-Ming Cheung, \textit{Fellow, IEEE} \and
Juyong Jiang}
\authorrunning{W.C. Lan et al.}
\institute{Department of Computer Science, Hong Kong Baptist University,\\ Hong Kong SAR, China\\
\email{\{cswclan, ymc, csjyjiang\}@comp.hkbu.edu.hk}}
% \end{comment}
%******************
\maketitle

\begin{abstract}
% Network pruning has been widely used and shown to be effective for compressing and accelerating deep neural networks. Existing pruning methods either removed weights and neurons or other integrated components such as channels and filters. However, the former 
Unstructured pruning has the limitation of dealing with the sparse and irregular weights. By contrast, structured pruning can help eliminate this drawback but it requires complex criterion to determine which components to be pruned. To this end, this paper presents a new method termed TissueNet, which directly constructs compact neural networks with fewer weight parameters by independently stacking designed basic units, without requiring additional judgement criteria anymore. Given the basic units of various architectures, they are combined and stacked in a certain form to build up compact neural networks. We formulate TissueNet in diverse popular backbones for comparison with the state-of-the-art pruning methods on different benchmark datasets. Moreover, two new metrics are proposed to evaluate compression performance. Experiment results show that TissueNet can achieve comparable classification accuracy while saving up to around 80\% FLOPs and 89.7\% parameters. That is, stacking basic units provides a new promising way for network compression. 
\end{abstract}

\section{Introduction}
Deep neural networks (DNNs) have obtained superior performance and become indispensable tools in computer vision community, such as image classification \cite{krizhevsky2012imagenet,he2016deep}, object detection \cite{girshick2014rich,he2017mask}, and semantic segmentation \cite{long2015fully,chen2017deeplab}. However, the widely-recognized properties of over-parameterization and redundancy result in huge consumption of memory footprint and computation cost, which hinder their real-life applications. Considering the limited storage space and computation capacity of mobile and edge devices that are often resource-constraint, it is desired to reduce the number of parameters and floating-point operations (FLOPs) of DNNs for better deployment \cite{cheng2018recent}. To obtain more efficient models, many techniques have been explored for compression and acceleration, including pruning \cite{han2015deep,ullrich2017soft}, quantization \cite{jacob2018quantization}, filter decomposition, knowledge distillation \cite{jin2019knowledge,romero2014fitnets}, and compact model design \cite{iandola2016squeezenet,howard2017mobilenets}. Among them, network pruning has been a mainstream branch in both academia and industry, showing broad prospects in numerous applications \cite{cheng2018model}. 

The pruning frameworks aim at eliminating the redundancy of deep models by removing the components with less importance. According to the types of removed components, there are mainly two kinds of pruning methods: unstructured pruning and structured pruning. Unstructured pruning removes the connections (weights) or neurons in weight matrices. A common standard to determine which weights should be pruned is magnitude-based pruning that compares the weight amplitude with a threshold \cite{han2015learning}.
% Firstly, a predefined quality parameter is multiplied by the standard deviation of a layer weight to calculate the threshold, and then the weight with lower magnitude than the threshold will be set to zero. After all layers are pruned, the model needs to be retrained so that the remaining weights can be adjusted to compensate for the removed ones. Since unstructured pruning has no structural restrictions on the position of the weights, it can prune the model without great damage of accuracy. 
However, this kind of low-level pruning has the risk of being non-structural that may hinder the actual acceleration, because of the irregular memory access mode. Some special software and hardware can alleviate this problem, but also brings extra cost to the deployment of models. 

To achieve more practical compression and acceleration, the well-supported structured pruning methods such as channel and filter pruning \cite{luo2017thinet,zhao2019variational,he2019filter} have been explored. Although the structured pruning for convolution kernels and graph can obtain hardware friendly network, they usually need specific criteria or complex mechanisms to identify the irrelevant subset of the components for deletion, which will increase the memory requirement and computation cost. For example, the reconstruction-based methods try to minimize the reconstruction error of feature maps based on the pretrained model to achieve pruning \cite{he2017channel,li2016pruning,luo2018thinet}, consuming much memory to store the feature maps of pretrained models. 

% On the other hand, most of the current methods on compression and acceleration are conducted on pretrained neural networks. For example, pruning methods try to remove the unimportant components of pretrained models and usually requires extra iterations to recover the accuracy; network quantization achieves compression through quantizing the parameters such as weights and activations in existing networks; low-rank decomposition decomposes the original weight tensors and approximates them such as using the multiplication of several low-rank matrices. 

Under the circumstances, it is interesting to ask whether it is possible to have a human-designed compact networks that can achieve comparable or even better compression performance than pruned networks learned through training. Since the current pruning methods are mainly conducted on pretrained neural networks, the performance of compressed model is highly affected by the quality of the pretrained models, which will lead to a restriction if the pretrained models are not well trained. Another concern is that network pruning often involves a lot of fine-tuning processes that may become an obstacle to implementation because the generalization ability of a given pruning framework in different architectures is usually unclear. Therefore, simply using a more efficient architecture may be more effective than pruning a suboptimal network in many cases \cite{blalock2020state}.

\begin{figure*}[t]
\centering
\includegraphics[width=1\columnwidth]{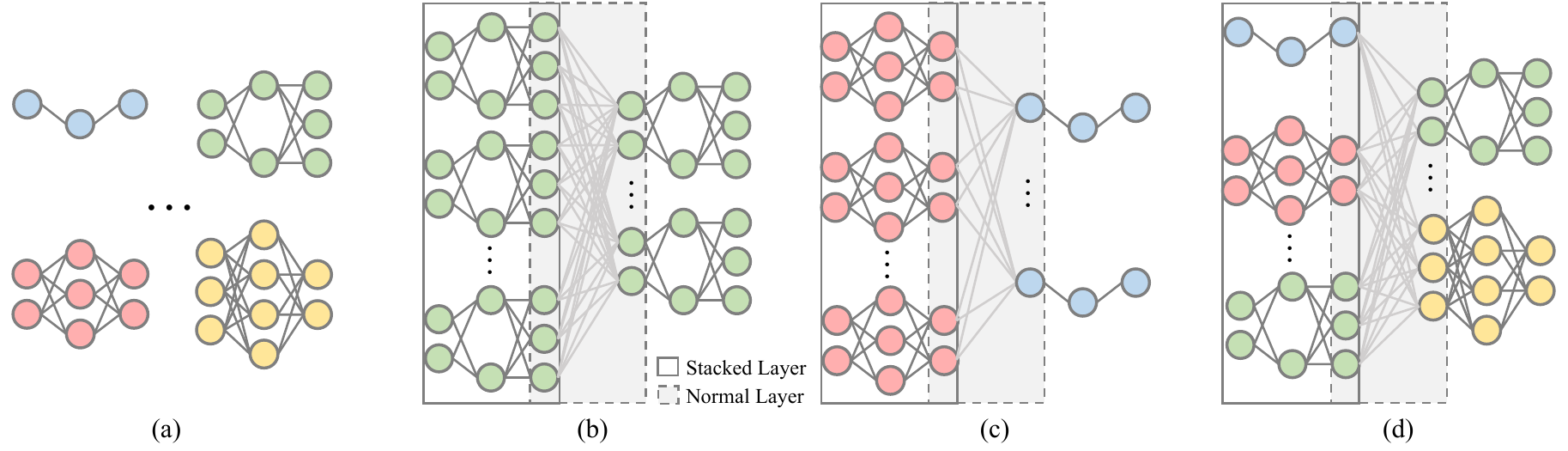}
\caption{Compact TissueNet with a combination of basic units. (a) Different basic units of various architectures. (b) TissueNet with a single kind of basic units. (c)-(d) TissueNet with hybrid kinds of basic units.}
\label{fig:tissuenet}
\end{figure*}

To this end, we propose TissueNet from a new perspective that directly constructs compact neural network by independently stacking designed basic units. We first design the basic unit as a small neural network with at least one hidden layer, and then we combine these basic units following a certain stacking strategy to construct an entire network. The whole framework of our proposed method is illustrated in Figure \ref{fig:tissuenet}. In the framework, a group of nodes (neurons) forms a basic unit just like the nervous tissues in human brain. Since there is no connection between each unit due to the independence, our TissueNet simply achieves weight compression without additional mechanisms for judging importance of weights. Different from the sparse and irregular weight tensors generated by unstructured pruning, the weights of each unit in TissueNet are regular and dense so that they can be easily processed without special devices. Besides, TissueNet can be built as numerous network backbones such as convolutional neural networks (CNNs) and full-connected networks by designing different basic units, with strong generalization ability. 

We conduct extensive experiments on multiple datasets (i.e., MNIST \cite{lecun1998mnist}, CIFAR \cite{krizhevsky2009learning}, Tiny ImageNet \cite{hansen2015tiny} and ImageNet-2012 \cite{deng2009imagenet}) with popular network structures (i.e., MLP, VGG \cite{simonyan2014very}, ResNet \cite{he2016deep}, and MobileNetV2 \cite{sandler2018mobilenetv2}). We also define new indicators for comparison with the state-of-the-art pruning methods. Experimental results demonstrate the efficacy of TissueNet for network compression on all datasets. In addition, we explore different combinations of various basic units on MNIST through ablation studies. Instead of relying on pretrained networks, we show that TissueNet can still work effectively compared with other pruning frameworks. This is a brand new attempt to construct compact networks under human-designed stacking structure, showing great potential on compressing network.

% parameters and FLOPs with acceptable accuracy decreases. 
% We summarize our main contributions as follows:
% \begin{itemize}
%     \item [1)]  We propose TissueNet that directly constructs compact neural networks by combining and stacking carefully designed basic units. It achieves compression due to the independence of each unit. Using different kinds of basic units, TissueNet can be applied on various models. 
%     \item [2)] Different basic units and combining strategy are explored to verify the flexibility and efficacy of TissueNet.
%     \item [3)] We empirically build TissueNet in different style on various widely-used backbones. The evaluations are conducted in image classification tasks with five popular datasets. We also define new indicators for comparison with other methods. The extensive experiments demonstrate that proposed TissueNet can obtain better compression performance. 
% \end{itemize}

\section{Related Work}
\subsection{Network Pruning}
Network pruning has been widely used to reduce the complexity and computation of CNNs because of its effectiveness and simplicity. Existing pruning method can be roughly divided into unstructured pruning and structure pruning in terms of the types of network components to be pruned. Unstructured pruning involves removing weights or neurons, while structured pruning includes removing the entire channels or filters. Unstructured pruning increases the sparsity of the network by replacing the connections or neurons with zero in the weight matrix. Han \emph{et al.} \cite{han2015deep} described a three-stage deep compression framework including network pruning, quantization and Huffman coding, where the weights were pruned by comparing the magnitude with a preset threshold. However, in such pruning technique, a weight will remain zero in the subsequent retraining process once it is removed which may cause large accuracy loss. Guo \emph{et al.} \cite{guo2016dynamic} then proposed a dynamic network pruning framework, introducing a recovery operation in order to restore the connection that has been wrongly pruned. 

Structured pruning breaks the limitation that unstructured pruning heavily relies on technical hardware or software library to achieve compression and speedup because of the sparse weights. ThiNet \cite{luo2017thinet} utilizes the features of the next layer to guide the pruning of the current layer. Other channel pruning methods have developed different strategies to choose the unimportant channels such as variational technique \cite{zhao2019variational} and genetic algorithm \cite{hu2018novel}. Filter-level pruning is also proved to be efficacious. Li \emph{et al.} \cite{li2016pruning} first calculated the norms of filters for evaluation. Later, He \emph{et al.} \cite{he2019filter} proposed to rank the importance of convolution filters based on geometric median to overcome the shortcoming of norm-based pruning. Parameter sharing is explored in \cite{zhang2018learning} to reduce model size according to the similarity between filters. Recent work \cite{sui2021chip} has introduced channel independence to measure the importance. Metapruning method \cite{liu2019metapruning} also combines Neural Architecture Search (NAS) \cite{yang2020cars} with pruning in these years. 

\subsection{Compact Model Design}
Compact model design methods focus on changing the basic operation and redesigning the structure so as to reduce the complexity and amount of parameters. It is a widely used method to construct a lightweight network with different cheaper operations. In the NIN \cite{lin2013network} model, the architecture of embedded network is developed that uses 1 × 1 convolution to increase capacity while decrease computational complexity. Besides 1 × 1 convolution, SqueezeNet \cite{iandola2016squeezenet} utilizes group convolution for further speedup. 
% The average pooling layer is used in GoogleNet \cite{szegedy2015going} to reduce parameter amounts. 
Howard \emph{et al.} \cite{howard2017mobilenets} designed MobileNet using branching strategy, where each branch contained only one channel called depth-wise convolution. The further work called MobileNetv2 \cite{sandler2018mobilenetv2} introduces residual and linear bottleneck structure. ShuffleNet \cite{zhang2018shufflenet} combines group convolution and channel shuffle operation that shuffles channels before the next convolution operation in order to realize information transmission between multiple groups. Furthermore, EspNetv2 \cite{mehta2019espnetv2} proposes deep expandable and separable convolution to improve efficiency. Jeon \emph{et al.} \cite{jeon2018constructing} proposed active shift operation to save memory, and then the sparse shift layer (SSL) applied in FE-Net \cite{chen2019all} eliminates the meaningless shift operation. Based on NAS \cite{yang2020cars} that can construct network automatically, \cite{tan2019mnasnet} and \cite{wu2019fbnet} have improved MobileNetV2 with comparable performance, but these methods still requires large resource consumption that is difficult to deploy on real-life applications.

% \begin{figure}[t]
% \centering
% \includegraphics[width=0.65\columnwidth]{Figure/tissuenet.png}
% \caption{Compact neural network with different basic units. }
% \label{fig:cellnet}
% \end{figure}

\section{Proposed Method}
\subsection{Preliminary}
Before describing our proposed method, we formally give some notations and symbols. In a CNN, the weight of a standard convolutional layer is a four-dimensional tensor $\mathbf{W} \in \mathbb{R}^{d \times d \times c_{in} \times c_{out}}$ , where $d \times d$ is the kernel size of the convolution filter, $c_{in}$ and $c_{out}$ is the number of input and output channels respectively. For a three-dimensional input $\mathbf{X}_{in} \in \mathbb{R}^{w_{in} \times h_{in} \times c_{in}}$ with width $w_{in}$ and height $h_{in}$, the convolutional layer can transform it into another three-dimensional tensor $\mathbf{X}_{out} \in \mathbb{R}^{w_{out} \times h_{out} \times c_{out}}$. Let Conv($\cdot$) denote the operations in a convolutional layer including convolution and activation, the transformation can be formulated as:
\begin{equation}\label{eq:generaloutput}
    \mathbf{X}_{out}= \text{Conv}(\mathbf{X}_{in},\mathbf{W}).
\end{equation}

For a given convolutional layer, the weight $\mathbf{W}$ contains $\{d \times d \times c_{in} \times c_{out}\}$ parameters, and the convolution operation requires $\{d \times d \times c_{in} \times w_{out} \times h_{out} \times c_{out}\}$ FLOPs. Since there are often a large number of convolutional layers in a CNN model, the cost of memory and computation will be huge that makes it a challenge to deploy CNNs on resource-constraint devices.

\subsection{Basic Unit and Stacked Layers}

As illustrated in Figure \ref{fig:tissuenet}, one of the most important components of the proposed TissueNet is the designed basic unit or called tissue. We take CNN as an example to demonstrate our method.
% To ensure that the unit has the ability to deal with non-linear problem, 
Without loss of generality, we initially design the unit as a small convolutional neural network with at least one hidden layer, where the nodes can be adjusted flexibly. Suppose the unit contains three layer and the node numbers in each layer is $\{c_{in}^{'},c_h,c_{out}^{'}\}$, respectively. The two weight matrices in a unit can be represented as $\mathbf{W}_l \in \mathbb{R}^{d \times d \times c_{in}^{'} \times c_{h}}$ and $\mathbf{W}_r \in \mathbb{R}^{d \times d \times c_{h} \times c_{out}^{'}}$. Then, for a given input $\mathbf{X}_{in}^{'} \in \mathbb{R}^{w_{in} \times h_{in} \times c_{in}^{'}}$, the output $\mathbf{X}_{out}^{'} \in \mathbb{R}^{w_{out} \times h_{out} \times c_{out}^{'}}$ of a unit can be calculated as:
\begin{equation}\label{eq:unitoutput}
    \mathbf{X}_{out}^{'} = \text{Unit} (\mathbf{X}_{in}^{'})= \text{Conv}(\text{Conv}(\mathbf{X}_{in}^{'},\mathbf{W}_l),\mathbf{W}_r).
\end{equation}

After designing the basic unit, we can then obtain the whole convolutional layers by stacking a certain number of units together. Suppose $m$ units with various $c_{in}^{'}$ are selected to stack and they are represented as $\{\mathbf{U}^1, \mathbf{U}^2, \dots, \mathbf{U}^m\}$. For the entire input $\mathbf{X}_{in} \in \mathbb{R}^{w_{in} \times h_{in} \times c_{in}}$ of the whole layer, it first needs to be split into $m$ pieces, where the input channel $c_{in}^{'}$ of each piece is the same as the corresponding unit, so that it can be processed with the unit because the unit only contains $c_{in}^{'}$ input channels. Note that the total input channels of $m$ units should be consistent with $c_{in}$.
If these $m$ units have the same input channel numbers, we can simple decide $m$ as $m=\frac{c_{in}}{c_{in}^{'}}$. 
Let $\{\mathbf{X}^1, \mathbf{X}^2, \dots, \mathbf{X}^m\}$ donate $m$ smaller input pieces with the corresponding $m$ outputs $\{\mathbf{X}^{1}_{out}, \mathbf{X}^{2}_{out},\dots,\mathbf{X}^{m}_{out}\}$. To ensure that the output of stacked layers can be transformed successfully into the next layer, these $m$ small outputs are also needed to be concatenated to keep the dimensions consistent. Thus, the output of the whole stacked convolutional layers is:
\begin{equation}\label{eq:stackedoutput}
 \begin{split}
    \mathbf{X_{out\_new}} & = \text{Concat}(\mathbf{X}_{in})\\
    &= [\mathbf{X}^{1}_{out}, \mathbf{X}^{2}_{out},\dots,\mathbf{X}^{m}_{out}] \\
    &= [\text{Unit}(\mathbf{X^1}),\text{Unit}(\mathbf{X^2}),\dots,\text{Unit}(\mathbf{X^m})],
 \end{split}
\end{equation}
where the output of each unit is calculated based on Equation \ref{eq:unitoutput}. 

By utilizing the simple stacking strategy, we can avoid the sparse weights generated after unstructured pruning. The weight tensors in TissueNet will remain \emph{dense} so that it can achieve compression using general-purpose hardware and software, without the support of complex judgment criteria as well. The detailed analysis on memory and computation cost is presented in Section \ref{sec:analysis}. In addition to CNN, TissueNet is also applicable to other network backbones by designing diverse basic units, as shown in Figure \ref{fig:tissuenet}(a).

\subsection{Network Construction}

In TissueNet, there are two kinds of convolutional layers termed as \emph{Stacked Layer} (solid box) and \emph{Normal Layer} (dashed box) as marked in Figure \ref{fig:tissuenet}. Stacked layer refers to the layers generated by stacking basic units while normal layer is the standard convolutional layer with full connection. Due to the independence of each unit in stacked layers, it is necessary to add normal convolution layers between the stacked layers; otherwise, the entire network will be disconnected resulting in the failure of feature information transmission. After obtaining the stacked layers as described in the previous section, a whole neural network can be constructed using these stacked layers and normal layers alternately. 

\begin{figure*}[t]
\centering
\includegraphics[width=0.9\columnwidth]{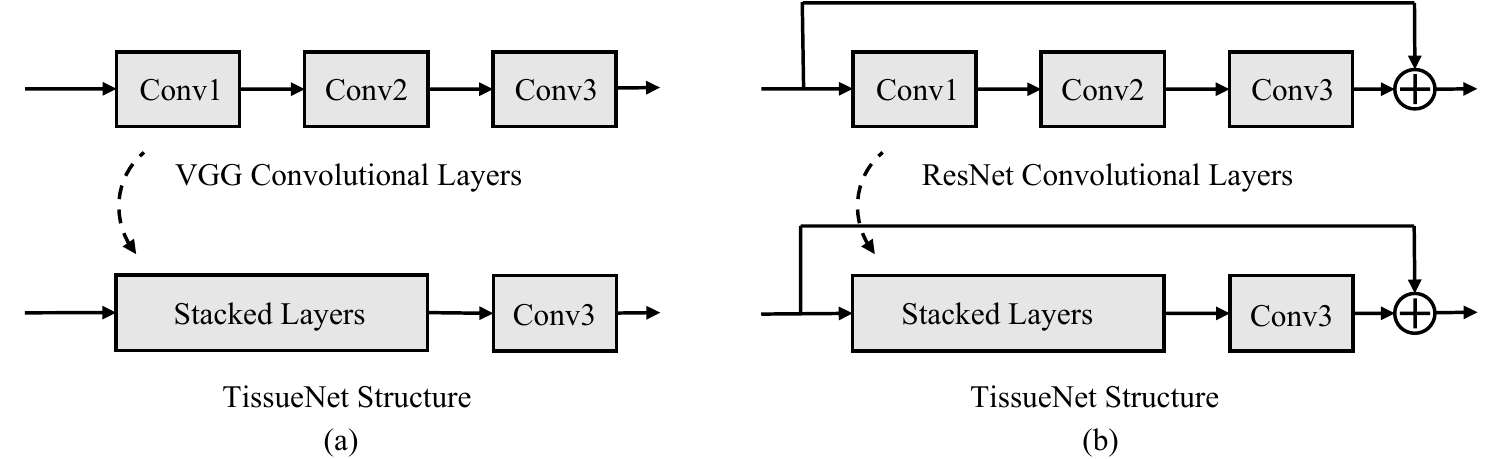}
\caption{Network construction. (a): VGG-style Network; (b): ResNet-style Network.}
\label{network}
\end{figure*}

It is straightforward to implement our proposed TissueNet with stacked layer as classical network architectures like VGG-style and ResNet-style backbones (style means the similar architecture). As shown in Figure \ref{network}(a), we can reconfigure the continuous standard convolutional layers in VGG-style networks by replacing the first few layers with stacked layers. With respect to ResNet-style in Figure \ref{network}(b), we can also introduce the residual structure in TissueNet. Furthermore, the constraint on dimensions between different layers can be flexibly solved by adjusting the parameters in stacked layer such as convolution stride. In addition to CNN, TissueNet is appropriate for full-connected networks like Multilayer Perceptron (MLP) if the unit is designed as a small network with only full-connected layers. In terms of the special convolution operations in recent lightweight models such as $3\times3$ depth-wise and $1 \times 1$ point-wise convolution, the basic units can also be designed to equip with these operations, demonstrating the powerful flexibility of the proposed TissueNet.

% \begin{figure}[t]\centering
% \subfigure[VGG-style Network.]
% {\centering \label{fig:vggstyle}
% \includegraphics[width=0.4\columnwidth]{Figure/network-a.pdf}}
% \subfigure[ResNet-style Network.]
% {\centering \label{fig:resnetstyle}
% \includegraphics[width=0.4\columnwidth]{Figure/network-b.pdf}}
% \caption{Network construction.}
% \label{network}
% \end{figure}

\subsection{Training Process}
TissueNet involves two parts of weights, one part is the weights of basic units in stacked layers $\{\mathbf{W}_l$, $\mathbf{W}_r$\} and the other is the weights of normal layers $\mathbf{W}_n$. Because each unit in TissueNet is still a neural network, it follows the standard forward and backward propagation algorithms. Given a set of data pairs $\{(\rm{\mathbf{x}},\rm{\mathbf{y})}\}$, the objective function of training TissueNet can be set as:
\begin{equation}\label{eq:traintissue}
\begin{split}
        {\arg\min}  \mathcal{L}(\mathbf{\hat Y}, \mathbf{y}) & = \underset{\mathbf{W,b}}{\arg\min} \mathcal{L}(f(C((\mathbf{W,b});\rm{\mathbf{x}}),\mathbf{y}))\\
    % & = \underset{\mathbf{W,b}}{\arg\min} \mathcal{L}(f(\mathbf{X}_{out}),\mathbf{y})
\end{split}
\end{equation}
where $\mathbf{W} = \{\mathbf{W}_l, \mathbf{W}_r,\mathbf{W}_n\}$ is all the weights and $b$ is the bias. The function $C(\cdot)$ contains the Conv operation of normal layers in Equation \ref{eq:generaloutput} and Concat operation in stacked layers as described in Equation \ref{eq:stackedoutput}. The function $f(\cdot)$ refers to the calculation in other layers such as fully-connection and pooling layers, and $\mathcal{L}(\cdot)$
is a selected convex loss function. 

When designing the basic unit (generally designed as three layers), the number of nodes in each of the layers $\{c_{in}^{'}, c_h, c_{out}^{'}\}$ needs to be preset. Then, the number of units $m$ and the input channels of $m$ input pieces can be adjusted based on each $c_{in}^{'}$. In forward propagation, the output of stacked layers is computed by Equation \ref{eq:stackedoutput} while the output of normal layers uses the standard convolution operation in Equation \ref{eq:generaloutput}. After obtaining the final output of TissueNet in forward propagation, we can get the gradient by performing backward propagation and update parameters using the popular optimizer such as SGD or ADAM \cite{kingma2014adam}. The training process of TissueNet is summarized in Algorithm \ref{alg:tissuenet}.

\begin{algorithm}[tb]
\caption{TissueNet Training}
\begin{algorithmic}
\State \textbf{Input}: Training data pairs $\{\mathbf{X}_{train}, \mathbf{y}_{train}\}$, number of nodes for base unit $\{c_{in}^{'},c_h,c_{out}^{'}\}$. 
\State \textbf{Output}: Compact neural networks.
\end{algorithmic}
\begin{algorithmic}[1]
    \State Get the number of basic units based on hyperparameter $c_{in}^{'}$ and stacked them.
    \State Initialize weight and bias in stacked layers and normal layers.
	\For{Iter = 1 to maxIter}
	  \State Get a minibatch of training data $\{\mathbf{x}, \mathbf{y}\}$.
	   \State Compute the network output where the output of stacked layers and normal layers is calculated as Equation \ref{eq:stackedoutput} and Equation \ref{eq:generaloutput}, respectively.
	\State Compute the loss $L(\mathbf{y}, \hat{\mathbf{y}})$.
	\State Perform standard backward propagation.
	\State Update parameters using any popular optimizer.
    \EndFor
\end{algorithmic}\label{alg:tissuenet}
\end{algorithm}

\subsection{Analysis of Stacked layers}\label{sec:analysis}

Suppose $m$ units with the same nodes $\{c_{in}^{'},c_h,c_{out}^{'}\}$ are stacked, a group of three convolutional layers will be constructed with $\{mc_{in}^{'},mc_h,mc_{out}^{'}\}$ nodes in each layer. Since the stacked units are independent from each other, which means that there is no connection or weight between them, the memory cost of stacked layers will be: 
\begin{equation}\label{stackmemory}
    M_s = m \mathbf{W}_l + m \mathbf{W}_r = m \times d \times d \times c_h \times (c_{in}^{'} + c_{out}^{'}).
\end{equation}
If three convolutional layers are fully connected, the two weight matrices will be $\mathbf{W}_{lf} \in \mathbb{R}^{d \times d \times mc_{in} \times mc_{h}}$ and $\mathbf{W}_{rf} \in \mathbb{R}^{d \times d \times mc_{h} \times mc_{out}}$ in which memory cost is: 
\begin{equation}\label{normalmemory}
   M_n = \mathbf{W}_{lf} + \mathbf{W}_{rf} = m^2 \times d \times d \times c_h  \times (c_{in} + c_{out}) = m M_s.
\end{equation}
Obviously, the number of parameters of stacked convolutional layers is much lower than the layers with fully connection. By this way, our proposed method can effectively reduce the memory cost to achieve networks compression. 

In terms of computation cost, the total FLOPs of $m$ three-layer units are:
\begin{equation}\label{stackflop}
C_s = m F_l + m F_r = m \times d \times d \times c_h \times( c_{in} \times w_{outl} \times h_{outl} +  w_{outr} \times h_{outr} \times c_{out}),
\end{equation}
where $w_{outl}$ and $w_{outr}$ are the width of the first and second layer output, respectively, $h_{outl}$ and $h_{outr}$ are the heights. For the full-connected convolutional layers with the same nodes, the FLOPs are:
\begin{equation}\label{normalflop}
C_n = d \times d \times mc_{h} \times (mc_{in} \times w_{outl} \times h_{outl} + w_{outr} \times h_{outr} \times mc_{out}) = mC_s.
\end{equation}
Similarly, the memory and computation cost in other architectures can also be calculated in the same way.

With less parameters and FLOPs, the stacked layers in our proposed TissueNet can help save much storage and computation consumption. Since we also add normal layers between the stacked layers and different kinds of units may be used when building network, the compression and acceleration of the entire TissueNet will be smaller than $m$. The actual compression performance of TissueNet will be reported in our experiments for comparison. Besides, hybrid kinds of units can also be applied to improve the generalization ability of TissueNet.

\section{Experiments}
To evaluate the performance of TissueNet, we conduct empirical experiments on five popular datasets for image classification tasks, that are MNIST \cite{lecun1998mnist}, CIFAR-10, CIFAR-100 \cite{krizhevsky2009learning}, Tiny ImageNet \cite{hansen2015tiny} and ImageNet-2012 \cite{deng2009imagenet}. We first construct TissueNet as full-connected MLP on MNIST, and then apply representative CNNs like  VGG \cite{simonyan2014very}, ResNet \cite{he2016deep}, and MobileNet \cite{sandler2018mobilenetv2}. We mainly compare TissueNet with baseline and the following pruning methods: % including channel and filter pruning:

\begin{itemize}%[leftmargin=*]
\item VP\cite{zhao2019variational}: Channel-level pruning using variational technique. 
\item NCP \cite{hu2018novel}: Channel-level pruning based on genetic algorithm.
\item DMCP \cite{xu2021efficient}: A mask-based channel pruning method.
\item DCP \cite{liu2021discrimination}: Channel-level pruning utilizing discrimination-aware losses. 
\item DI-unif \cite{hou2021discriminant}: Pruning channels according to discriminant information.
\item PFEC \cite{li2016pruning}: Pruning filters via weight magnitudes.
\item CHIP \cite{sui2021chip}: Filter pruning through channel independence.

\end{itemize}

\subsection{Performance Score}
The most commonly used metrics to assess the performance of a compressed model are numbers of parameters and FLOPs. However, when considering the \emph{trade-off} between compression and accuracy, it will be unfeasible to evaluate different models. Thus, we define new unified indicators to measure the compressed model termed \emph{Computation Efficiency} (CE) and \emph{Storage Efficiency} (SE). CE is related to FLOPs calculating as the contribution of a unit FLOPs makes to accuracy compared with baseline, that is $CE = (acc_n/flops_n) / (acc_b/flops_b)$, where $flops_n$ and $flops_b$ is the FLOPs of new compressed model and baseline model respectively, while $acc_n$ and $acc_b$ are the accuracy. Similarly, SE is the contribution of a unit parameters makes to accuracy compared with baseline, computing as $SE = (acc_n/param_n) / (acc_b/param_b)$. From the definition, the two scores of baseline will be 1 and higher scores means higher resource utilization. Using CE and SE, we can fairly compare the performance of different compressed models.

\subsection{Experiment Setting}
\subsubsection{TissueNet Construction.} 
We build the diverse kinds of TissueNet under different network architectures to verify the generalization ability. For the three hyperparamters $\{c_{in}^{'}, c_h, c_{out}^{'}\}$ in basic unit of MLP-style and VGG-style network, we fix $c_{in} = c_{out} = 2$ and only adjust $c_h$ to evaluate the performance. Following the reconfiguration in Figure \ref{network}, we replace the layers in original VGG and ResNet with our stacked layers and treat the original networks as baseline. 
To simplify the structure, we replace the last fully-connection layers with average pooling layer. 
When constructing the stacked layers in TissueNet, we consider both single and hybrid kinds of units. On ResNet-style network, we explore different strategies to deploy stacked layers, such as replacing the intermediate blocks while remaining the layers in the first and last blocks. 

\subsubsection{Training Strategy.}
For CIFAR datasets, we train TissueNet for 500 epochs with the initial learning rate of 0.1 and batch size of 128. TissueNet on MNIST and Tiny-ImageNet is trained 200 epochs and other setting is the same as CIFAR. Stochastic gradient decent (SGD) with momentum 0.9 is used as optimizer on both CIFAR and Tiny-ImageNet. For ImageNet, the batch size is 128 as well. The initial learning rate is set as 0.01 adjusted with the cosine annealing during 200 training epochs. Adam with $10^{-4}$ weight decay and 0.01 epsilon is applied as optimizer. The cross-entropy loss is adopted as a criterion for all datasets. In addition, all the training procedures are warmed up with 10 epochs and applied weight initialization. As for the network backbones, VGG-style TissueNet is trained on CIFAR and Tiny-ImageNet, while ResNet-style TissueNet is applied on CIFAR and ImageNet. Besides, we also evaluate the performance of lightweight model MobileNetV2 on CIFAR.

\subsection{MNIST}
In Table \ref{tissuenetmnist}, we report the performance of TissueNet on MNIST using MLP. We choose the MLP with two hidden layers as baseline, where the number of nodes in each layer is "784-500-300-10" as applied in \cite{chen2020dynamical}. The hypermeter $c_h$ in our method is set to 4. In addition to the widely used metrics like accuracy, number of FLOPs and parameters, we also report the two proposed scores SE and CE for comparison with other methods. It can be seen that, we can achieve 97\% FLOPS and 96\% parameters reduction of baseline model, with only 0.7\% accuracy loss. Besides, we can also obtain much higher CE and SE scores compared with the other three pruning methods. If we apply larger $c_h$, the accuracy will also be improved slightly. Thus, it is shown that our proposed TissueNet can work well on full-connected neural networks.

\begin{table*}[t]
\small
\centering
\caption{Results on MNIST.}
\centering
\begin{tabular}{ccccccc}
% \hline
\bottomrule[0.7pt]
Model & Method & Acc(\%) & FLOPs (M) & Param (M) & CE & SE\\
\hline
\multirow{5}*{\textbf{MLP}} & Baseline & 92.65 & 1.09 & 0.55 &1 &1\\
& TissueNet($c_h$ = 4) & 91.95 & 0.03 & 0.02 & \textbf{\textit{36.06}} & \textbf{\textit{27.29}} \\
& SSL \cite{wen2016learning} & 92.53 & 0.17 & 0.09 & 6.40 & 6.10  \\
& NS \cite{liu2017learning} & 92.49 & 0.17 & 0.09  & 6.40 & 6.10 \\
& CAC \cite{chen2020dynamical} & 92.72 & 0.17 & 0.09 & 6.42 & 6.12 \\
\bottomrule[0.7pt]
\end{tabular}
\label{tissuenetmnist}
\end{table*}

\subsection{CIFAR-100 and CIFAR-10}

\subsubsection{CIFAR-100.} We first apply VGG-style TissueNet with different $c_h$ in Table \ref{tissuenetcifar100}. When $c_h$ is set as 4, we obtain an accuracy of 70.40\% with 1.57G FLOPs and 4.08M parameters. It means that we manage to prune around 80\% FLOPs and 73\% parameters of the baseline model. By increasing $c_h$ to 6, the accuracy gap between TissueNet and baseline can be narrowed to 2.4\%. The results show that TissueNet can effectively achieve compression without great loss of accuracy. Although the accuracy of TissueNet is a bit lower than state-of-the-art methods such as NCP \cite{hu2018novel} whose accuracy is much close to baseline, we can get higher utilization ratio of unit FLOPs and parameters since the CE and SE scores are much larger. Besides, we can improve the accuracy by adjusting $c_h$ or even the layers of basic units. For ResNet-style, we try to replace the layers in ResNet-18 with our stacked layers in TissueNet except for the first and last blocks (represented as the symbol 'r'). Under this replacing strategy, we can prune 19\% FLOPs and 10\% parameters of baseline, where the accuracy loss is only 0.76\%. If we change all the block layers in ResNet-18 into stacked layers, the CE and SE scores can be improved which means better utilization of computation and storage resources. 

\subsubsection{CIFAR-10.} In Table \ref{tissuenetcifar10}, the proposed ResNet-style TissueNet can obtain a CE score of 1.59 that is much higher than the other state-of-the-art methods. It achieves 39\% pruning on FLOPs and 40\% parameters compared to baseline model, where the accuracy drop is acceptable. Applying different replacing strategy, the accuracy gap will be further narrowed. On VGG-style networks, TissueNet still has better performance on computation efficiency with only 0.43\% decrease on accuracy and the FLOPs can be saved for 79.87\%. Though the SE score of CHIP \cite{sui2021chip} is larger than TissueNet, the accuracy loss is higher at the same time.
On the whole, the results on CIFAR dataset verify the flexibility and effectiveness of the proposed TissueNet on network compression.

\begin{table*}[!t]
\small
\centering
\caption{Results on CIFAR-100.}\centering
\begin{tabular}{ccccccc}
% \hline
\bottomrule[0.7pt]
% \hline
Model & Method & Acc(\%) & FLOPs ($\times 10^8$) & Param (M) & CE & SE\\
\hline
% \multicolumn{7}{c}{\textbf{CIFAR-100}}\\
% \hline

\multirow{7}*{\textbf{VGG-16}} & Baseline & 73.55 & 7.79 & 14.96 &1 &1\\
&TissueNet($c_h$ = 6) & 71.13 & 1.67 & 4.14 & 4.51 & 3.49\\
&TissueNet($c_h$ = 4) & 70.40 & 1.57 & 4.08 & \textbf{\textit{4.74}} & \textbf{\textit{3.51}}\\
& VP \cite{zhao2019variational}& 73.33 & 6.39 & 9.29 & 1.21 & 1.61\\
& NCP \cite{hu2018novel} & 73.35 & 4.88 & 5.27 & 1.59 & 2.83\\
& DMCP\cite{xu2021efficient} & - & - & - & 1.05 & 1.41\\
& PFEC \cite{li2016pruning} & - & - & - & 1.03 & 1.20 \\
\hline
\multirow{6}*{\textbf{ResNet-18}} & Baseline & 77.32 & 11.13  & 11.22 &1 &1\\
&TissueNet($c_h$ = 4,'r') & 76.56 &  9.03 & 10.16 & 1.22 & 1.09\\
&TissueNet($c_h$ = 4) & 75.07 & 4.52 & 4.20 & \textbf{\textit{2.39}} & \textbf{\textit{2.60}} \\
& TAS \cite{dong2019network} & -& - & -& 2.23 &-\\
& FPGM \cite{he2019filter} & -& - & -& 2.35 &-\\
& DI-unif \cite{hou2021discriminant} & -& - & -&  2.21 &\textbf{\textit{2.73}}\\

\hline
\multirow{4}*{\textbf{MobileNetV2}} & Baseline & 72.58 & 1.35 & 2.37 &1 &1\\
&TissueNet('r') & 70.36 & 0.68 & 1.78  &  1.92 & 1.29  \\
% & LeGR \cite{chin2019legr}  & -& - & -&  &-\\
& DI-unif \cite{hou2021discriminant} & 68.95 & 0.68& -& 1.90  &-\\
\bottomrule[0.7pt]
\end{tabular}
\label{tissuenetcifar100}
\end{table*}

\begin{table*}[!t]
\small
\centering
\caption{Results on CIFAR-10.}\centering
\begin{tabular}{ccccccc}
% \hline
\bottomrule[0.7pt]
% \hline
Model & Method & Acc(\%) & FLOPs ($\times 10^8$) & Param (M) & CE & SE\\
\hline
\multirow{8}*{\textbf{VGG-16}} & Baseline & 93.25&	7.80&	14.91 &1 &1\\
&TissueNet($c_h$ = 4)&	92.82&	1.57&	4.04&	\textbf{\textit{4.95}}&	3.67\\
&VP\cite{zhao2019variational}	& 93.18	& 4.75 & 3.97&	1.64	&3.75\\
& NCP \cite{hu2018novel}	& 93.28	&-	&-	&2.27	&6.25\\
& DMCP \cite{xu2021efficient}	&-	&-	&-	& 1.33	& 3.30\\
& PFEC \cite{li2016pruning}	&93.40	&5.13	&5.4	&1.52	&2.77\\
& CHIP \cite{sui2021chip}	&92.47	&1.67	&1.89&	4.63	&\textbf{\textit{7.82}}\\

% & [6]PFEC & - & - & - & 1.15 & 1.67 \\
\hline
\multirow{5}*{\textbf{ResNet-18}} & Baseline & 95.29 & 11.13 & 11.17 &1 &1\\
&TissueNet($c_h$ = 6,'r') & 94.50 & 9.04 & 10.11 & 1.22 & 1.10 \\
&TissueNet($c_h$ = 6) & 93.31 & 6.84 & 6.65 & \textbf{\textit{1.59}} & 1.64 \\
& DMCP \cite{xu2021efficient} & -&-  &-  & 1.54 & \textbf{\textit{5.70}}  \\
& PFEC \cite{li2016pruning} & - & - & - & 1.11 & 1.49 \\

\hline
\multirow{5}*{\textbf{MobileNetV2}} & Baseline & 92.20 & 1.35 & 2.25 &1 &1\\
&TissueNet('r') & 91.14 & 0.97 & 1.67  & \textbf{\textit{1.37}} & \textbf{\textit{1.33}} \\
& WM \cite{zhuang2018discrimination} & - & - & - & 1.35 & 1.30 \\
% & NPPM \cite{gao2021network} & - & - & - & 1.90 & - \\
& DCP \cite{liu2021discrimination} &  - & - & - & 1.36 & 1.30\\
% \hline
% \hline
\bottomrule[0.7pt]
\end{tabular}
\label{tissuenetcifar10}
\end{table*}

% \begin{figure}[t]
% \centering
% \subfigure[Test Accuracy on CIFAR-100 using ResNet-18.]{\centering \label{fig:c100convergence}\includegraphics[width=0.35\columnwidth]{Figure/acc_epoch_a.pdf}}
% \hspace{8mm}
% \subfigure[Test Accuracy on CIFAR-10 using VGG-16.]{\centering \label{fig:c10convergence}\includegraphics[width=0.35\columnwidth]{Figure/acc_epoch_b.pdf}}
% \caption{Convergence Curve on CIFAR dataset.}
% \label{fig:convergence-curve}
% \end{figure}

\begin{figure*}[t]
\centering
\includegraphics[width=1\columnwidth]{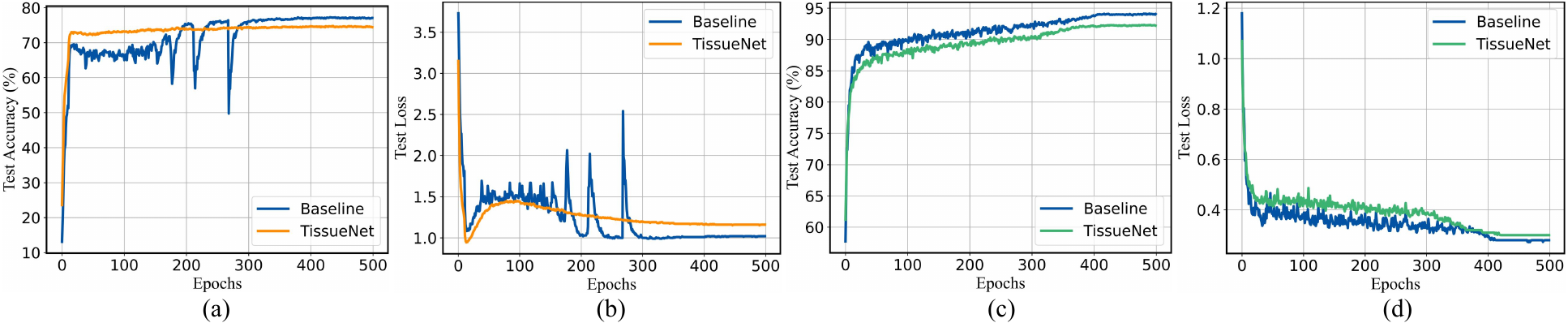}
\caption{Convergence Curve on CIFAR dataset. (a)/(b): Test Accuracy/Loss on CIFAR-100 using ResNet-18; (c)/(d): Test Accuracy/Loss on CIFAR-10 using VGG-16.}
\label{fig:convergence-curve}
\end{figure*}

\subsubsection{Lightweight Model.} In additional to classical CNNs, we also evaluate TissueNet on lightweight model MobileNetV2 that is also popular in recent years. In MobileNet, there exist special 1 $\times$ 1 pointwise and 3 $\times$ 3 depthwise convolution operations, where the group number in depthwise convolution equals to the number of input channels. Our TissueNet can easily deal with these special operations by adjusting the parameters in basic units such as the kernel size and group number. The results are summarized in Tables \ref{tissuenetcifar100} and \ref{tissuenetcifar10}. It can be seen that our TissueNet can also be implemented successfully as lightweight models. For instance, we get an accuracy of 91.14\% on CIFAR-10, although there is a 1.1\% loss compared to baseline, it is encouraging that the FLOPs and parameters can be pruned by 28\% and 26\% respectively. Moreover, the utilization ratio of FLOPs and parameters are both somewhat higher than other methods. On CIFAR-100, the accuracy gap between TissueNet and baseline is relatively large. However, it manages to save 50\% FLOPs and 25\% parameters, and the performance score of CE is still larger than DI-unif \cite{hou2021discriminant}. Overall, the results demonstrate that TissueNet can work efficiently on lightweight model.

\subsubsection{Convergence.} On CIFAR dataset, we check the convergence of TissueNet as illustrated in Figure \ref{fig:convergence-curve}. Figure \ref{fig:convergence-curve}(a)(b) displays the convergence curve on CIFAR-100 using ResNet-18. It is worth noting that TissueNet can almost reach an optimal status after certain training epochs. Besides, the training process of TissueNet is more stable though the maximum accuracy is a bit lower than baseline. On CIAFR-10 uisng VGG-16 in Figure \ref{fig:convergence-curve}(c)(d), our TissueNet is trained to be converged in fewer epochs compared with original VGG network, providing the potential possibility to acceleration.

\subsection{Tiny-ImageNet and ImageNet}

\begin{table*}[tb]
\small
\centering
\caption{Results on Tiny-ImageNet and ImageNet.}\centering
\begin{tabular}{ccccccc}
\toprule[0.7pt]
% \hline
Model & Method & Top-1 Acc(\%) & FLOPs (G) & Param (M) & CE & SE\\
\hline
\multicolumn{7}{c}{\textbf{(a) Tiny-ImageNet}}\\
\hline
\multirow{5}*{\textbf{VGG-19}} & Baseline & 55.65	& 3.80 &	20.63 & 1 & 1\\
&TissueNet($c_h$ = 6) &  50.20 & 0.70 & 4.72 & \textbf{\textit{4.90}} & \textbf{\textit{3.94}} \\
&TissueNet($c_h$ = 6,'r') &  53.54 & 1.39 & 9.36 & \textbf{\textit{2.84}} & \textbf{\textit{2.29}} \\
& DMCP \cite{xu2021efficient} & -&-  &-  & 1.21 & 2.02 \\
& PFEC \cite{li2016pruning} & - & - & - & 1.05 & 1.36 \\
\hline
\multicolumn{7}{c}{\textbf{(b) ImageNet}}\\
\hline
\multirow{5}*{\textbf{ResNet-34}} & Baseline & 73.31	& 113.82 & 21.82 &1 &1\\
& TissueNet($c_h$ = 8,'r') & 71.85 & 80.72 & 17.53 &\textbf{\textit{1.38}} & 1.22\\
& PFEC \cite{li2016pruning} & 72.20 & - & 19.30 & 1.30 & 1.11 \\
& Taylor-FO \cite{molchanov2019importance} & 72.83 & - & 17.20 & 1.28 & 1.26 \\
& NISP \cite{yu2018nisp} & 73.03 & - & - & 1.37 & \textbf{\textit{1.37}} \\
\bottomrule[0.7pt]
\end{tabular}
\label{tissuenetimage}
\end{table*}

\subsubsection{Tiny-ImageNet.} We apply VGG-19 on Tiny-ImageNet and the results are reported in Table \ref{tissuenetimage}(a). Compared with baseline, the accuracy drop of TissueNet is relatively large if we replace all the layers with stacked layers, but it can save more than 81.5\% FLOPs and 77\% parameters. As analysed in previous section, we can improve the accuracy by adjusting the hyperparameters to add nodes or applying different replacing strategy. When we use the same strategy as ResNet that only replacing the intermediate layers, the accuracy can be improved to 53.54\% that is comparable with baseline. Compared with other pruning methods, our TissueNet can still obtain a relatively larger score both on CE and SE, showing the efficacy again.

\subsubsection{ImageNet.} On large-scale ImageNet, we select ResNet-34 as the baseline model and compare the performance of our TissueNet with state-of-the-art pruning methods in Table \ref{tissuenetimage}(b). It is encouraging that our method achieves 71.85\% top-1 accuracy and 29\% FLOPs degradation, with higher CE score than other methods. Also, we manage to reduce the amount of parameters by around 20\%. To further increase the accuracy, hybrid kinds of units can be selected to construct network. These results demonstrate that the proposed TissueNet is efficient to reduce computation and memory consumption on complex tasks, meanwhile maintaining a comparable accuracy. 

% \begin{figure}[t]
% \centering
% \subfigure[]{\centering \label{fig:c100flops}\includegraphics[width=0.24\columnwidth]{Figure/res-c100-flop.PNG}}
% \subfigure[]{\centering \label{fig:c100param}\includegraphics[width=0.245\columnwidth]{Figure/res-c100-param.PNG}}
% \subfigure[]{\centering \label{fig:c10flops}\includegraphics[width=0.24\columnwidth]{Figure/res-c10-flop.PNG}}
% \subfigure[]{\centering \label{fig:c10param}\includegraphics[width=0.245\columnwidth]{Figure/res-c10-param.PNG}}
% \caption{Trade-off curve of TissueNet on CIFAR-100 and CIFAR-10 using ResNet-18. (a)/(b): Number of FLOPs/parameters on CIFAR-100; (c)/(d): Number of FLOPs/parameters on CIFAR-10. }
% \label{fig:tradeoff-curve}
% \end{figure}

\begin{figure*}[t]
\centering
\includegraphics[width=1\columnwidth]{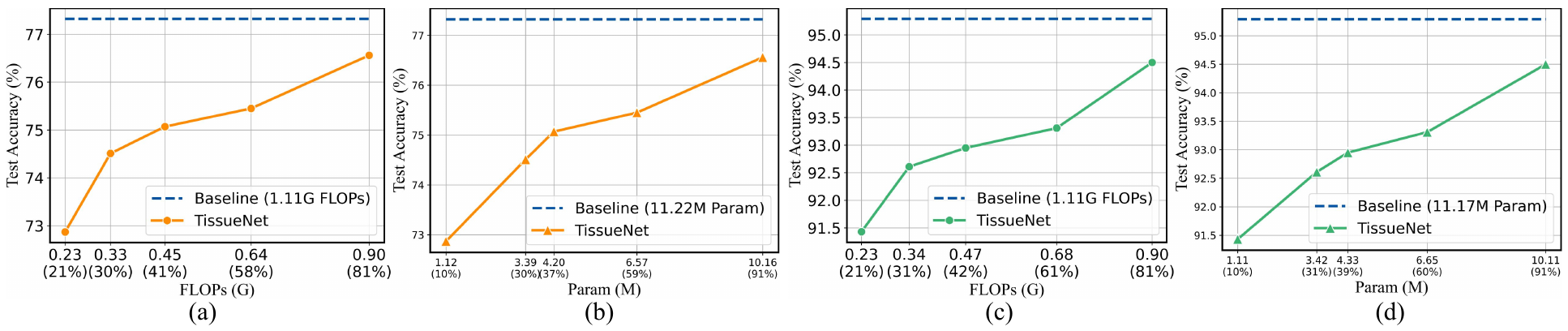}
\caption{Trade-off curve of TissueNet on CIFAR-100 and CIFAR-10 using ResNet-18. (a)/(b): Number of FLOPs/parameters on CIFAR-100; (c)/(d): Number of FLOPs/parameters on CIFAR-10.}
\label{fig:tradeoff-curve}
\end{figure*}

\subsection{Ablations}
\subsubsection{Varying FLOPs and Parameters.}

To study the performance under different compression ratios, we apply various $c_h$ and replace strategies to adjust FLOPs and parameters for compact ResNet-style TissueNet on CIFAR-10 and CIFAR-100. The results in Figure \ref{fig:tradeoff-curve} help comprehensively understand the trade-off of the proposed TissueNet. The four curves reveal the same trend that TissueNet will perform worse with the decrease of number of FLOPs and parameters. This trend is also consistent with our experience that more compact model will sacrifice more accuracy. On CIFAR-100, when we tune the remaining FLOPs from 21\% to 33\% and corresponding parameters from 10\% to 30\%, the compression ratio drops a little but the accuracy increases drastically from 72.87\% to 74.51\%. The same observation is also verified on CIFAR-10. It is reasonable that continuing to compress will result in a sharp drop in accuracy after the model is compressed to a certain extent. Thus, to achieve a better trade-off between compression and accuracy, we empirically control the remaining FLOPs and parameters as more than 20\% except for MNIST in our experiments.

\subsubsection{Hybrid Units.}

We also apply hybrid kinds of units in stacked layers to verify the flexibility of TissueNet as illustrated in Figure \ref{fig:tissuenet}(d). Specifically, we design three different units $\{\mathbf{U}^1, \mathbf{U}^2, \mathbf{U}^3\}$ with increment on FLOPs and parameters, and construct TissueNet as the similar architecture of LeNet-5 \cite{lecun1998gradient}. The performance of TissueNet using single and hybrid units are reported in Table \ref{hybrid}. We first build the network separately with three kinds of units. For hybrid units, one of the three units is randomly selected to stack for each unit position in stacked layers. To exclude fortuity in random selection, we perform the evaluation using hybrid units for three times. Compared with single unit, hybrid units have the potential to help improve the accuracy, when maintaining the FLOPs and parameters as the same compression level. For example, we adjust the numbers of three units stacked in $\text{Hybrid}^2$ to keep the FLOPs and parameters same as using single unit $\mathbf{U}^2$, it is observed that hybrid units lead to a higher accuracy. The results on $\mathbf{U}^1$ and $\mathbf{U}^2$ also indicate the possibility that unit with less FLOPs and parameters can still bring better accuracy. Therefore, these observations motivate us to explore more effective units and hybrid combinations.

% \begin{table*}[tb]
% \small
% \centering
% \caption{Results on MNIST using hybrid units.}\centering
% \renewcommand\arraystretch{1.25}
% \begin{tabular}{cccccc}
% \toprule[0.7pt]
% Model & FLOPs(M) & Param(k)  & Acc(\%) & CE & SE \\
% \hline
% Baseline & 1.45 & 13.67 & 99.05 & 1 & 1\\
% $\mathbf{U}^1$ & 0.72 & 1.41 & 98.61 & 2.0 & 9.6 \\
% $\mathbf{U}^2$ & 0.95 & 1.83 & \textbf{98.31} & 1.51 & 7.41 \\
% $\mathbf{U}^3$ & 1.19 & 2.26 & 98.72 & 1.21 & 6.03 \\
% $\text{Hybrid}^1$ & 0.95 & 1.83 & \textbf{98.65} & 1.52 & 7.44 \\
% $\text{Hybrid}^2$ & 0.92 & 1.78 & 98.47 & 1.57 & 7.63 \\
% $\text{Hybrid}^3$ & 1.04 & 1.99 & 98.59 & 1.39 & 6.84\\
% \bottomrule[0.7pt]
% \end{tabular}
% \label{hybrid}
% \end{table*}

\begin{table*}[tb]
\small
\centering
\caption{Results on MNIST using hybrid units.}\centering
\renewcommand\arraystretch{1.15}
\begin{tabular}{cccc|cccc}
\toprule[0.7pt]
\multicolumn{4}{c|}{\text{Single}} & \multicolumn{4}{c}{\text{Hybrid}} \\
\hline
Model & FLOPs(M) & Param(k)  & Acc(\%) & Model & FLOPs(M) & Param(k)  & Acc(\%) \\
\hline
$\mathbf{U}^1$ & 0.72 & 1.41 & 98.61 & $\text{Hybrid}^1$ & 0.92 & 1.78 & 98.47 \\
$\mathbf{U}^2$ & 0.95 & 1.83 & 98.31 & $\text{Hybrid}^2$ & 0.95 & 1.83 & 98.65 \\
$\mathbf{U}^3$ & 1.19 & 2.26 & 98.72 & $\text{Hybrid}^3$ & 1.04 & 1.99 & 98.59 \\

\bottomrule[0.7pt]
\end{tabular}
\label{hybrid}
\end{table*}

\section{Conclusion}
This paper is the first attempt to construct human-designed compact networks called TissueNet, which combines a number of designed basic units to generate stacked layers. Compared with standard layers with fully connections, the stacked layers contain much less parameters and FLOPs due to the independence of each basic unit so that TissueNet can achieve compression. We have also provided analysis of the memory and computation cost of TissueNet in detail. The results show that the proposed TissueNet can obtain better performance than the other pruning methods, with a good trade-off between accuracy and compression ratio. This implies the flexibility and efficacy of human-designed compact networks, providing a new promising way for network compression.

% \newpage
\bibliographystyle{splncs04}
\bibliography{egbib}
\end{document}